\documentclass[conference]{IEEEtran}
\IEEEoverridecommandlockouts
\usepackage{cite}
\usepackage{amsmath,amssymb,amsfonts}
\usepackage{algorithmic}
\usepackage{graphicx}
\usepackage{textcomp}
\usepackage{xcolor}

\usepackage{times}  %
\usepackage{helvet} %
\usepackage{courier}  %
\usepackage[hyphens]{url}  %
\usepackage{graphicx} %
\usepackage{subcaption}
\usepackage{amsmath}
\usepackage{amssymb}
\usepackage{caption} %
\usepackage{algorithm}
\usepackage{algorithmic}

\usepackage{array}
\newcolumntype{P}[1]{>{\centering\arraybackslash}p{#1}}

\def\BibTeX{{\rm B\kern-.05em{\sc i\kern-.025em b}\kern-.08em
    T\kern-.1667em\lower.7ex\hbox{E}\kern-.125emX}}

\begin{document}

\makeatletter
\newcommand{\linebreakand}{%
  \end{@IEEEauthorhalign}
  \hfill\mbox{}\par
  \mbox{}\hfill\begin{@IEEEauthorhalign}
}
\makeatother

\title{
  Coinbot: Intelligent Robotic Coin Bag Manipulation Using Deep Reinforcement Learning And Machine Teaching\\
{}
\thanks{}
}

\author{\IEEEauthorblockN{Aleksei Gonnochenko}
\IEEEauthorblockA{\textit{Sber Robotic Center} \\
\textit{PJSC Sberbank}\\
Moscow, Russia \\
gonnochenko.a.s@sberbank.ru}
\and
\IEEEauthorblockN{Aleksandr Semochkin}
\IEEEauthorblockA{\textit{Sber Robotic Center} \\
\textit{PJSC Sberbank}\\
Moscow, Russia \\
semochkin.a.n@sberbank.ru}
\and
\IEEEauthorblockN{Dmitry Egorov}
\IEEEauthorblockA{\textit{Sber Robotic Center} \\
\textit{PJSC Sberbank}\\
Moscow, Russia \\
egorov.d.alek@sberbank.ru}
\linebreakand %
\IEEEauthorblockN{Dmitrii Statovoy}
\IEEEauthorblockA{\textit{Sber Robotic Center} \\
\textit{PJSC Sberbank}\\
Moscow, Russia \\
statovoy.d.a@sberbank.ru}
\and
\IEEEauthorblockN{Seyedhassan Zabihifar}
\IEEEauthorblockA{\textit{Sber Robotic Center} \\
\textit{PJSC Sberbank}\\
Moscow, Russia \\
zabikhifar.s@sberbank.ru}
\and
\IEEEauthorblockN{Aleksey Postnikov}
\IEEEauthorblockA{\textit{Sber Robotic Center} \\
\textit{PJSC Sberbank}\\
Moscow, Russia \\
postnikov.a.l@sberbank.ru}
\linebreakand %
\IEEEauthorblockN{Elena Seliverstova}
\IEEEauthorblockA{\textit{Sber Robotic Center} \\
\textit{PJSC Sberbank}\\
Moscow, Russia \\
seliverstova.e.vl@sberbank.ru}
\and
\IEEEauthorblockN{Ali Zaidi}
\IEEEauthorblockA{\textit{Microsoft} \\
Berkeley, CA \\
alizaidi@microsoft.com}
\and
\IEEEauthorblockN{Jayson Stemmler}
\IEEEauthorblockA{\textit{Neal Analytics} \\
Seattle, WA \\
jayson@nealanalytics.com}
\and
\IEEEauthorblockN{Kevin Limkrailassiri}
\IEEEauthorblockA{\textit{Neal Analytics} \\
Seattle, WA \\
kevinl@nealanalytics.com}
}

\maketitle

\begin{abstract}
Given the laborious difficulty of moving heavy bags of physical currency in the cash center of the bank, there is a large demand for training and deploying safe autonomous systems capable of conducting such tasks in a collaborative workspace. However, the deformable properties of the bag along with the large quantity of rigid-body coins contained within it, significantly increases the challenges of bag detection, grasping and manipulation by a robotic gripper and arm. In this paper, we apply deep reinforcement learning and machine learning techniques to the task of controlling a collaborative robot to automate the unloading of coin bags from a trolley. To accomplish the task-specific process of gripping flexible materials like coin bags where the center of the mass changes during manipulation, a special gripper was implemented in simulation and designed in physical hardware. Leveraging a depth camera and object detection using deep learning, a bag detection and pose estimation has been done for choosing the optimal point of grasping. An intelligent approach based on deep reinforcement learning has been introduced to propose the best configuration of the robot end-effector to maximize successful grasping. A boosted motion planning is utilized to increase the speed of motion planning during robot operation. Real-world trials with the proposed pipeline have demonstrated success rates over 96\% in a real-world setting.
\end{abstract}


\section{Introduction}

The developing direction of collaborative robots suggests that universal robot manipulators that can work safely alongside humans in existing workspaces, may partially or completely take over operations for manipulating objects in an autonomous fashion. This means that robots will be deployed in environments where a human has previously worked and will continue to work, and the environment should not require modifications with the addition of a robot. On the one hand, this is an inexpensive way to automate manual operations, and can have enormous business impact through the dramatic increase in productivity and efficiency in the workplace, allowing human operators to shift from laborious manual tasks to more cognitive heavy tasks. On the other hand, robot control systems deployed in such non-deterministic and safety critical environments are subject to increased safety and intelligence requirements than the usual industrial robots in production which act mostly independently of human operators. However, training robotic manipulators to use sensory input, like their human counterparts, to locate objects within their vicinity to grasp, and then plan and execute a safe grasp and manipulate operation is a very challenging problem. It requires coordination of vision, continuous control, and awareness of safety requirements. Sensor noise and occlusions can obscure the precise geometric position of objects in the environment. Moreover, the laws of  physics governing how mass and friction are neither observed directly by the robotic controller, nor learned intrinsically through evolutionary experience, as is the case for humans.

Deep reinforcement learning (DRL) has recently emerged as one of the most promising approaches for training autonomous systems in difficult non-deterministic environments. To learn suitable representations that can be utilized for control policies, DRL generally requires an exceptionally large number of training samples. Unfortunately, generating these samples in a real robotic systems is arduous because of the high sampling cost, and can be prohibitive due to safety precautions. Such a dilemma between the difficulty of feature extraction and high sampling cost is one of the most crucial issues for making the DRL approach more practical in real robotic control. In this paper, we present an application of the Bonsai Machine Teaching Platform for automating an operation previously performed by a human, and requiring minimal changes to the environment where the work is performed by a robot.

In this paper, we examine the workflow of a Bank's cash processing center, where a human operator maintains a coin counting machine. For processing, coins are transported in a mobile cart that is filled to the top with bags weighing up to 6 kg. Our goal is to develop approaches to enable a collaborative robot to identify and unload individual bags of coins from the cart and place them in a prescribed order on the table of the human operator.

The solution of the problem is complicated by the following environmental characteristics: bags have a variable shape and a movable center of gravity when manipulated, their weight varies from several tens of grams to 6 kg, the bags are located in carts with varying heights and depths where bags are frequently in positions that are difficult or impossible to reach by the manipulator in its current orientation, and bags can slip and even tear when the grip is unsuccessful. In all these situations, the robot control system should respond in a way that ensures continuous autonomous operation. %

There has been a surge of interest in applying recent advances in RL to the application of pick and place or assembly tasks \cite{Neef2020TowardsIP}, \cite{Mahler2017LearningDP}, \cite{Mahler2019LearningAR}, \cite{Breyer2019ComparingTS}. A reinforcement learning approach for policy learning involves training function approximators such as deep neural networks to predict the probability of success of candidate grasp orientation and positions from images based on empirical successes and failures through repeated active interactions with a simulation environment. To segment individual bags by visual perception, we utilize recent deep learning methods for image recognition \cite{wu2019detectron2}. To solve the problem of bag grasping, we have developed a gripper that can hold a heavy deformable object weighing up to 6 kg with an unstable center of gravity during manipulation and not break the shell of the bag. Since the robot operates mainly in a static environment when automating the bags unloading operation, we have proposed a method for speeding up motion planning. This method reduces the time of path planning when positioning the robot's end effector inside a deep trolley is performed.

\section{Related Work}

Robotic grasp detection is a challenging task, but the past few years have demonstrated that the application of deep learning architectures are capable of incredible feats in this domain. Moreover, the grasping of solid bodies has seen great success thanks to imitation learning \cite{Mahler2019LearningAR}, \cite{Mahler2017LearningDP}, ensembled Convolution Neural Nets \cite{Park2018RealTimeHA}, and standard reinforcement learning methods without geometric object models \cite{Gualtieri2018PickAP}. Some prior work has attempted to learn to complete a manipulation task without having to calculate paths and grasping points in a simulation setup using reinforcement learning \cite{Timmers2018LearningTG}, or has utilized large offline datasets for pre-training a generic policy using diverse experience replay data \cite{Kalashnikov2018QTOptSD}.

Compared to the rigid object manipulation, deformable object manipulation has additional challenges. For example shape control during grasp is major issue which was investigated in \cite{Nadon2019AutomaticSO}. In \cite{Saxena2019GarmentRA} authors focused on grasping point detection for deformable objects such as the garments. In \cite{Tsurumine2019DeepRL} presented two new deep reinforcement learning algorithms and conducted real robot experiments on a t-shirt folding task. In deployment-time for the real robot, they accelerated learning by initializing both presented networks with demonstration data, and calculated rewards from visual feedback. In our work, we introduce a novel approach to predict a proper configuration of the end-effector for successful grasping using curriculum learning applied to current state-of-the-art DRL algorithms. The proposed results of the paper take inspiration from prior work \cite{Corona2018ActiveGR} for target grasping point detection using deep learning, which has been demonstrated to be successful when applied to garments. While finding promising results in simulation, the pipeline presented in \cite{Corona2018ActiveGR} formulated the problem as a supervised learning problem, where the sole objective is gripper localization, and did not consider sequential decision making as part of the problem. In this work, we extend the consideration of not just gripper localization, but also gripper orientation, and frame the problem as a Markov Decision Process to account for the sequential nature of the task.

To increase the speed of robot operation, motion planning operations should aim to be as fast as possible. In this regard, we were motivated by recent results to develop a boosted motion planning. In \cite{Kim2020MotionPO} the motion planning of robot manipulators was designed using a Twin Delayed Deep Deterministic Policy Gradient (TD-DDPG) algorithm. They showed that a DRL approach can plan smoother and shorter paths than those designed by PRM.  In \cite{Rakita2018RelaxedIKRS} a real-time motion-synthesis method was presented for robot manipulators, called \textit{RelaxedIK}, is able to create smooth, feasible motions that avoid joint-space discontinuities, self-collisions, and kinematic singularities. In this paper, we proposed a new approach based on analytical approaches for constructing a boosted motion planning system because of the need to immerse the robot manipulator in the limited space of the trolley.

\subsection{Contributions}

In previous works, standard or customized RL approaches have been utilized and some of them have been tested only in simulation. In grasping point detection methods, they focus on locating the optimal position of grasping, but do not focus on orientation. To our knowledge, our work is the first where both the position and orientation of grasping point are jointly learned in the reinforcement learning policy. Moreover, this paper also conducts real experiments to demonstrate the successful sim-to-real transfer.

Building on top of previous work, the contributions of our paper are as follows:

\begin{enumerate}
    \item Formulates grasping and lifting of coinbags as a Markov Decision Process
    (MDP), modeling the process of grasping and removing coinbags from
    a flat trolley surface, and placing them onto an adjacent table.
    \item A new application for collaborative-robots: cash center currency processing for a major commercial bank
    \item A highly optimized new boosted motion planner to speed up the robot operation
    \item A fine-tuned perception module for extracting point-clouds of the
    bag positions, requiring very few samples of actual bag positions.
    \item A complete simulation environment in MuJoCo for the patented gripper and the robotic arm
    \item A newly developed and patented gripper for grasping the deformable coin bags while the center of mass is changing during manipulation
\end{enumerate}

\section{Methodology and Approach}

In this section, we will discuss the framework encompassing the robotics operations for unloading bags from a trolley. Since we formulated our problem as a reinforcement learning problem for optimizing the configuration of the robot end effector for successful grasping, the structure of the final solution is centered around an intelligent agent that receives state data from the environment and sends the recommended actions back to the environment, as shown in Figure \ref{interface}.

\begin{figure}
    \center
    \includegraphics[scale=0.22]{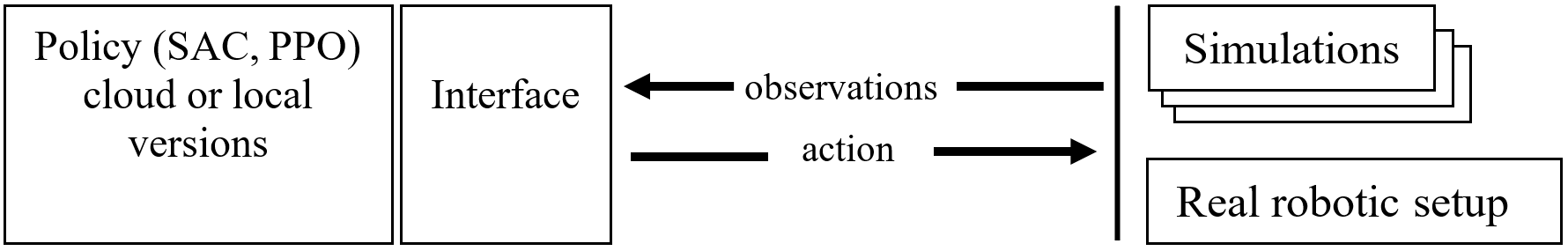}
    \caption{A common interface capable both simulation and real-life robot control}
    \label{interface}
\end{figure}

We utilize a variety of mechanisms to reduce training costs. First, we separate perception and control by fine-tuning a pre-trained object detection and instance segmentation model on a small dataset of coinbags (100 images in total). Second, following the methodology of curriculum learning \cite{Bengio2009CurriculumL}, \cite{Portelas2019TeacherAF}, we guide the training of our models by progressively increasing the complexity of the training environment with the agent’s performance. Finally, the entire training is performed in simulation and we report the required adjustments and findings from transferring policies to a real robotic arm and gripper.

\begin{figure}
    \center
    \includegraphics[scale=0.225]{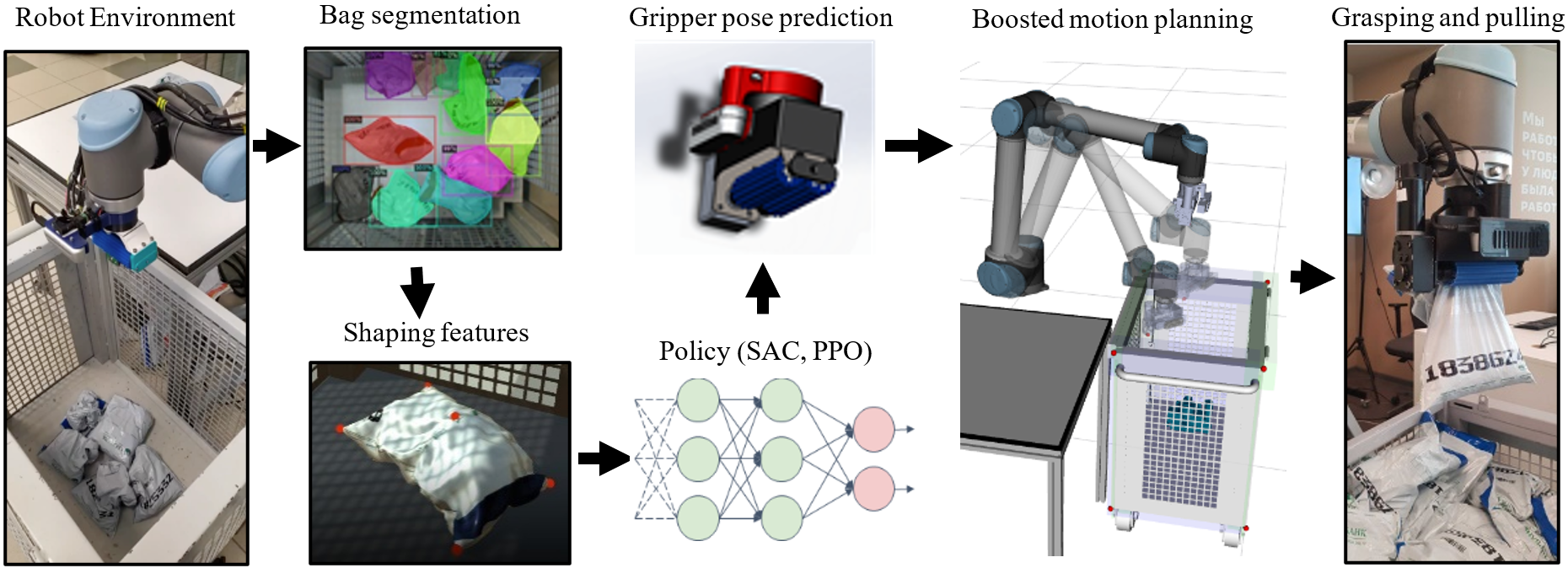}
    \caption{Robotic Setup for Training and Deployment}
    \label{robotic-setup}
\end{figure}

Figure \ref{robotic-setup} shows a diagram of the decision-making pipeline from the perception of the working area by the robot to the grasping of the bag. The information passed to the agent about the state of the environment includes 3D coordinates of multiple bag coordinates and details its location in the trolley, which are determined based on the point clouds of the bag that is captured via a camera. The procedure of getting this information in the simulation and in the real installation differs only at the stage of obtaining point clouds. In a real trials, bag segmentation occurs in the RGB image retrieved by camera and outputs the corresponding point cloud based on RGBD data. In the simulation, the point cloud is calculated from the structure of the 3D bag object. Further calculations of the input information for the agent are identical, therefore it is possible to use an interface that allows to interact uniformly with both simulators and the real-world installation.

At output, the trained intelligent agent provides recommendations for position and orientation of the gripper. The next important element of the proposed solution is the generation of the trajectory for a six-axis manipulator with a gripper at the end. In the case we are considering, the gripper must be positioned in a deep trolley, which imposes additional requirements on the trajectory calculation time. To quickly build trajectories under static scene constraints, we used a boosted motion planning model, described in a subsequent section.

\subsection{Robotics setup}

Following a thorough profiling and analysis of the various procedures in cash centers, it was determined that the most time-consuming procedure in the cash centers was the unloading of bags of coins on the table before registration. The weight of one bag is up to 6 kilograms, and the weight of bags in one basket can reach 500 kilograms. Therefore, in one working day, a single operator of the sorting machine is responsible of the transfer of a few tons of coin bags without assistance.

\begin{figure}[htbp]
    \centering
    \includegraphics [width=0.47\textwidth]{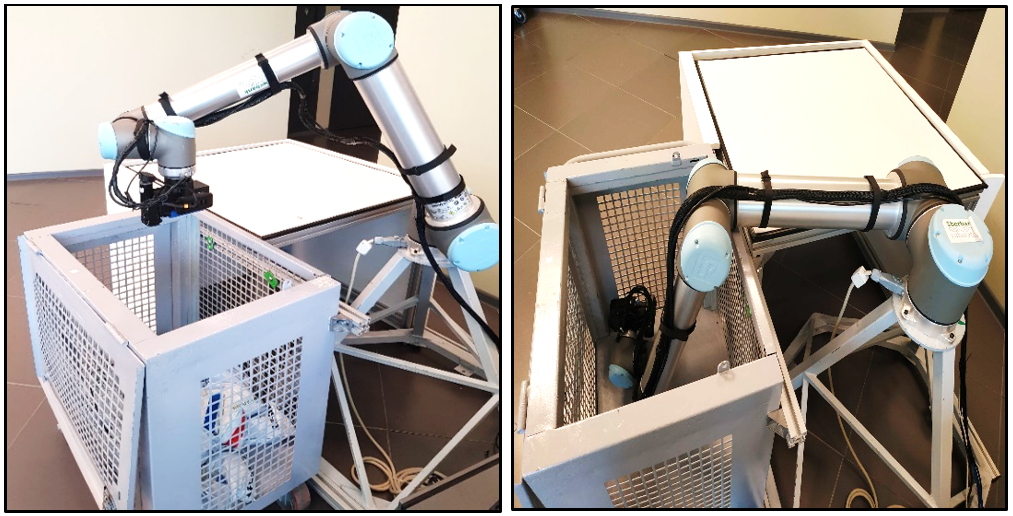}
    \caption{Robotic interaction with trolley}
    \label{end-effector}
\end{figure}

Due to constraints on manipulator length and number of degrees of freedom, we should properly choose relative position of the trolley to the table and the robot. We have tested the scene configuration manually in simulations to check reachability of regions inside the trolley. This allows us to verify that robot manipulator will be able to grasp all the bags with different poses. The right panel of Figure \ref{end-effector} shows a particular pose of gripper, where the collision margin is relatively small.

A collaborative manipulator UR10 was used in this project, and the manipulator has six degrees of freedom. The collaborative property of the UR10 arm allows it to be used safely in collaboration alongside the human operator. The carrying capacity of this robot allows us to carry a bag of up to 6 kilograms. We applied two control modes for robot movement – normal mode and force mode. The normal mode has been used to perform long motions between trolley and the table, and also allows the robot to move close to the coin bag inside the trolley before the grasping operation. The force mode has been used for grasping operation and allows the manipulator to make a compliant motion along the selected axis in the robot's workspace. Therefore, when the robotic arm contacts solid objects, such as bags of coins, in this mode, the robot will stop smoothly.

We used a real trolley, which is already in use at the banking cash center. The trolley is a metal box with 4 caster wheels at the bottom. Its walls are perforated, which complicates the task of detecting bags inside the trolley due to complex spotted shadows. Dimensions of the trolley are 500x700x800 millimeters. In banking cash center, trolleys can move freely relative to the table but for the reason described above in our work, we assumed that the trolley is fixed relative to the table and the robot and its position and orientation are known.

The table dimensions are 880x880x770 millimeters, and is set as a destination for bags where 8 areas are marked for bag placement. The bag of coins which is used in banking cash center is a soft plastic (5-layer nylon co-ex film) container. It has good barrier properties against gas, moisture and another destructive influence. Dimensions of the bag without coins are 290x180 millimeters.

\subsection{Gripper Design}

There are many types of grippers presented in the literature. In general, they are segmented into two types: gripper with suction cups and two finger or multi-finger grippers. Although grippers are widely used in soft body manipulation, we present our new patented gripper \cite{efimov_gonnochenko} for use in this project, which is specialized for deformable bags which contain rigid-body objects.

\begin{figure}
    \centering
    \includegraphics[width=0.95\linewidth]{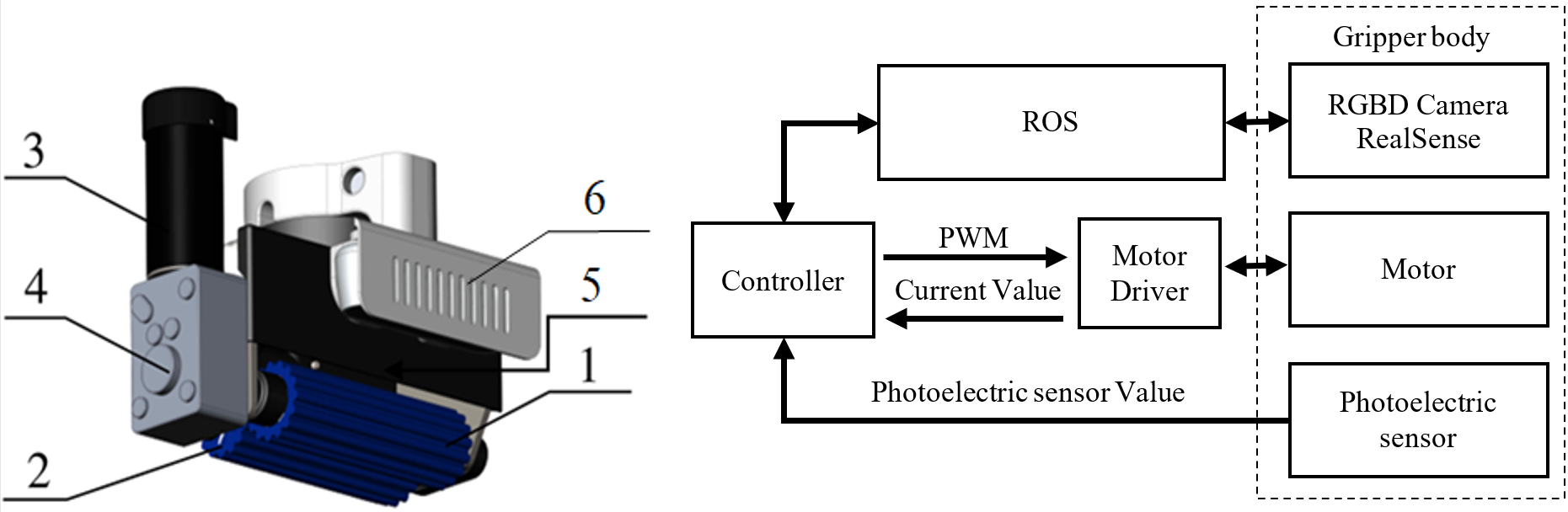}
    \caption{Gripper Design}
    \label{fig:gripper-design}
\end{figure}

The end effector (Figure \ref{fig:gripper-design}) consists of two mounting rollers (1, 2), the electric motor and gearbox (3 , 4), infrared sensor E3Z-D61 (5) and RGBD RealSense camera D435 (6). Rollers covered by polymer material with high friction coefficient which is proper for deformable bag grasping.

The principle of operation of the gripper is that the motor rotates the drive roller (1) around its axis clockwise and the driven roller (2) is rotated counterclockwise. Such rotation of the rollers during grasping the bag, provides the tightening of the part of the elastic bag due to the frictional force between the rollers and the bag. The motor stops when it is sufficient to pull the bag and it is tightened between the gripper rollers. For this, the current feedback is checking when the grasping process starts and a photoelectric sensor inside the box of the gripper confirms successful grasp. The operation of releasing the grasping bag is carried out by rotating the rollers in the opposite direction from the initial one.

\begin{figure}
    \centering
    \includegraphics[width=\linewidth]{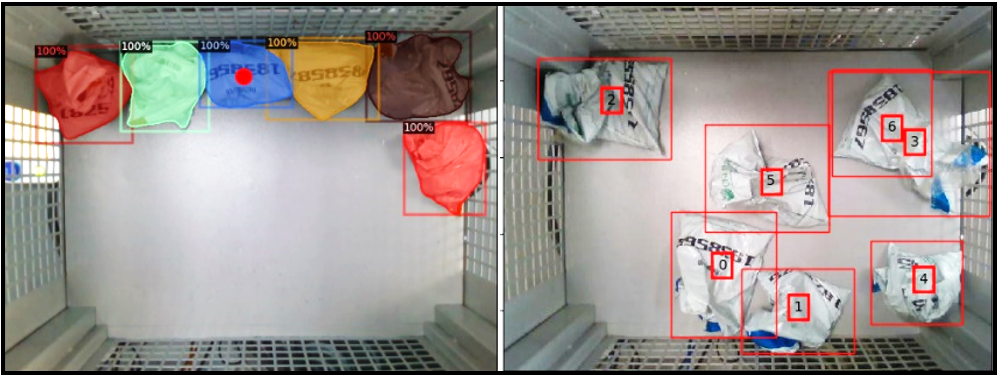}
    \caption{Bag Detection, Segmentation and Ranking}
    \label{fig:bag-perception}
\end{figure}

\subsection{Perception}

For recognizing bags and extracting bag point clouds and coordinates, a fine-trained Detectron2 model was utilized, due its training and inference speed, \cite{wu2019detectron2}. To create a labelled dataset for fine-tuning an existing Detector2 model (pre-trained on ImageNet), we utilized a collection of 100 images of bags in the trolley  at different positions and heights. For the segmentation of bags, an RGB image is fed to the detector input. In the output, we get a set of coordinates of $R_j$ pixels that are included in the selected areas. In Figure \ref{fig:bag-perception} segmented objects and ordering the priority for grasping have been shown.

Since operations with point clouds significantly increase processing time, we first select one bag after the segmentation operation, and only calculate its point cloud to get obtain the state input for the agent. The set of the areas are sorted in descending order of the area value, then the first three areas are sorted by closest distance from the center of the RGB image. The Figure \ref{fig:bag-perception}, right side, shows a possible fault in bag segmentation. Therefore, areas that intersect with other areas more than 80\% of the area are excluded from the list. From the list sorted in this way, the first $R_{s}$ element is selected, which is a candidate for extraction by the robot.

For each point $(u_{i},v_{i})\in R_{s}$ of the depth image using the camera intrinsic matrix, the corresponding points $(x_{i},y_{i},z_{i})$ on the surface of a real object in three-dimensional space are calculated. Therefore, for the selected bag, we have a corresponding cloud $C_{s}$ with 3D points, $N=\left|C_{s}\right|$. To simplify the perception model, we have presented each bag with 5 primary coordinates: the 4 corners of the bag and the center of the bag. This simplified representation increases the training time in simulation and operating time in real experiments. Let's denote $P_{i}=(a_{0}^{i},a_{1}^{i},a_{2}^{i})$ the i-th point of the cloud $C_{s}$ and let $(P_{i})_{(i=1)}^{N}$ be a sequence of points $P_{i}$.

\[
S_{k}\cdot\left(P_{i}\right)_{i=1}^{N}\overset{\text{sorted by value }a_{k}^{i}}{\longrightarrow}\left(P_{i}^{\prime}\right)_{i=1}^{N}.
\]

The following approach is used to calculate the characteristic points
$\left(A_{i}\right)_{i=0}^{4}$ of the bag. $A_{0}$ is the first
element of the sorted sequence $S_{0}\left(S_{1}\left(\left(P_{i}\right)_{i=1}^{N}\right)\right)$
and $A_{1}$ is the last element. $A_{2}$ is the first element of
the sorted sequence $S_{1}\left(S_{0}\left(\left(P_{i}\right)_{i=1}^{N}\right)\right)$
and $A_{3}$ is the last element. If we denote $A_{i}=\left(a_{0}^{i},a_{1}^{i},a_{2}^{i}\right)$,
then
\begin{align*}
x_{0} & =\left(a_{0}^{2}+a_{0}^{3}\right)/2\\
y_{0} & =\left(a_{1}^{0}+a_{1}^{1}\right)/2
\end{align*}
The fifth characteristic point $A_{4}$ is found like a point from
$C_{s}$ with first two coordinates $(a_{0}^{i},a_{1}^{i})$ nearest
to the $(x_{0},y_{0})$. The resulting sequence $\left(A_{i}\right)_{i=1}^{4}$
is considered as the result of perception and characterizes the geometric
properties of the selected bag.

\section{Simulation and Inverse Kinematics}

Since coin bags are deformable and the center of gravity is variable, we have selected a simulator that will most plausibly simulate the dynamics of a filled soft body, as well as the interaction of the gripper device with such bags. Composite objects from MuJoCo 2.0 \cite{Todorov2012MuJoCoAP} are used to create, simulate and render complex deformable objects. After each response from the agent, the process of moving the gripper to the recommended position is simulated. Subsequently, the interaction of the gripper rollers with the bag surface is simulated. If the grabbing device successfully grasps the bag, then its movement to the point above the trolley is simulated. At this point, the bag could fall out of the gripper due to insufficient contact of the gripper rollers with the bag shell, or a sharp shift in the center of gravity of the bag. In this way, the agent learns to take into account the features of the bag structure and select the most convenient position of the gripper device next to the bag inside the cart.

Simulation of such complex objects as bags of coins in MuJoCo imposes excessive demands on computer performance.  Therefore, two simulators are utilized to improve simulation speed and decouple the separate tasks. The first simulator is for grasping simulation, in which interaction between gripper rollers and bag is simulated (Figure \ref{fig:simulation} a). The second simulator is for checking collisions between entire manipulator and environment objects (Figure \ref{fig:simulation} b) while calculating inverse kinematic task for pose of the gripper recommended by agent.

\begin{figure}
    \centering
    \begin{subfigure}{0.196\textwidth}
      \centering
      \includegraphics[width=\linewidth]{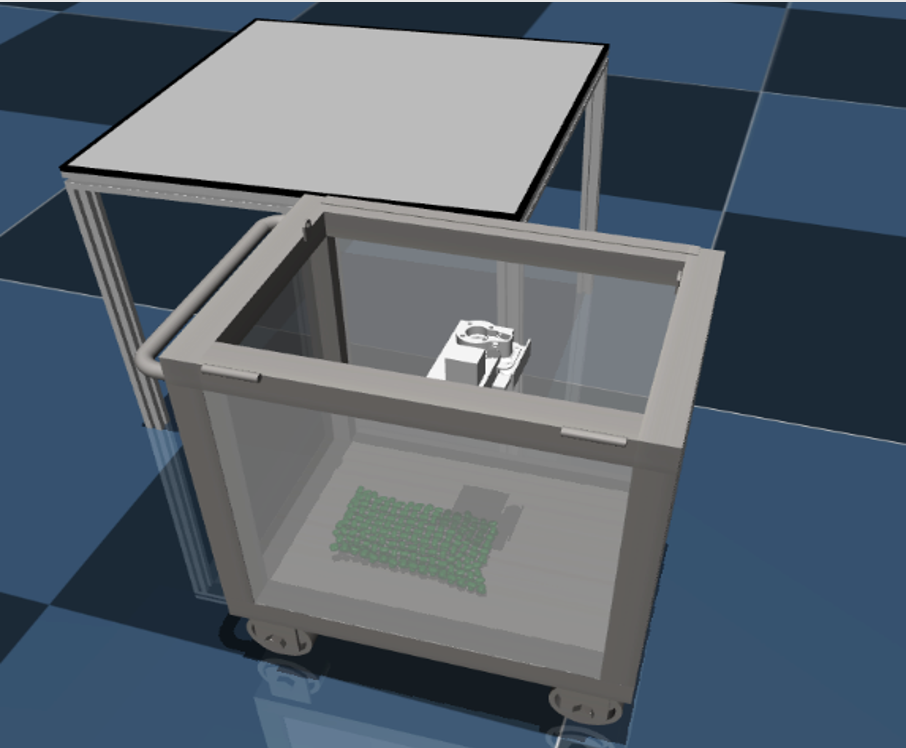}
      \caption{gripper simulation}
    \end{subfigure}%
    \begin{subfigure}{0.20\textwidth}
      \centering
      \includegraphics[width=\linewidth]{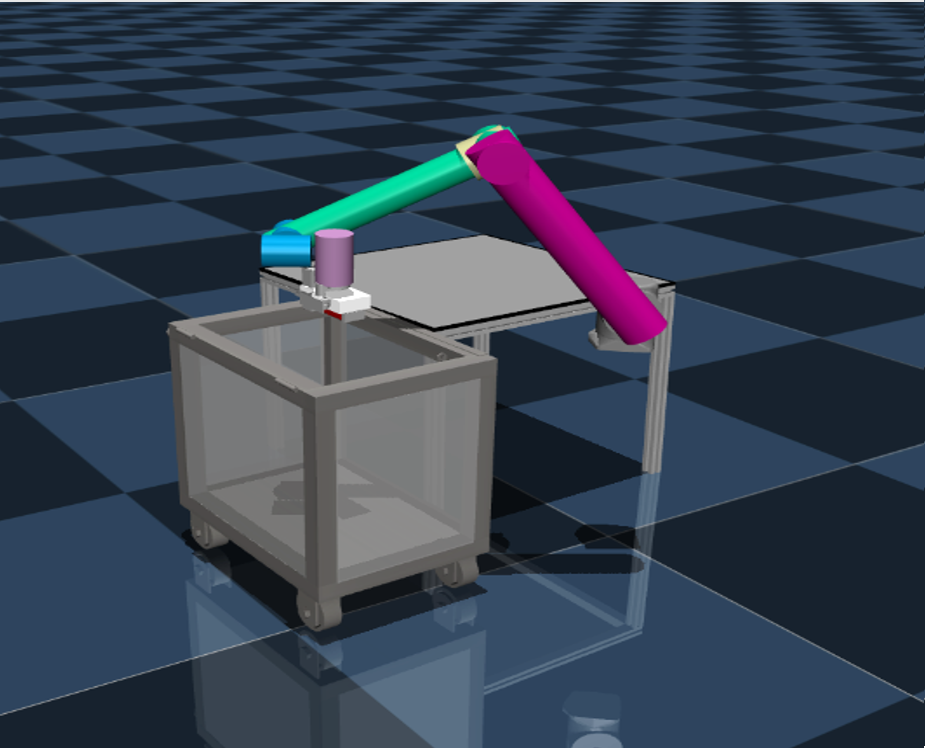}
      \caption{arm simulation}
    \end{subfigure}
    \begin{subfigure}{0.20\textwidth}
      \centering
      \includegraphics[width=\linewidth]{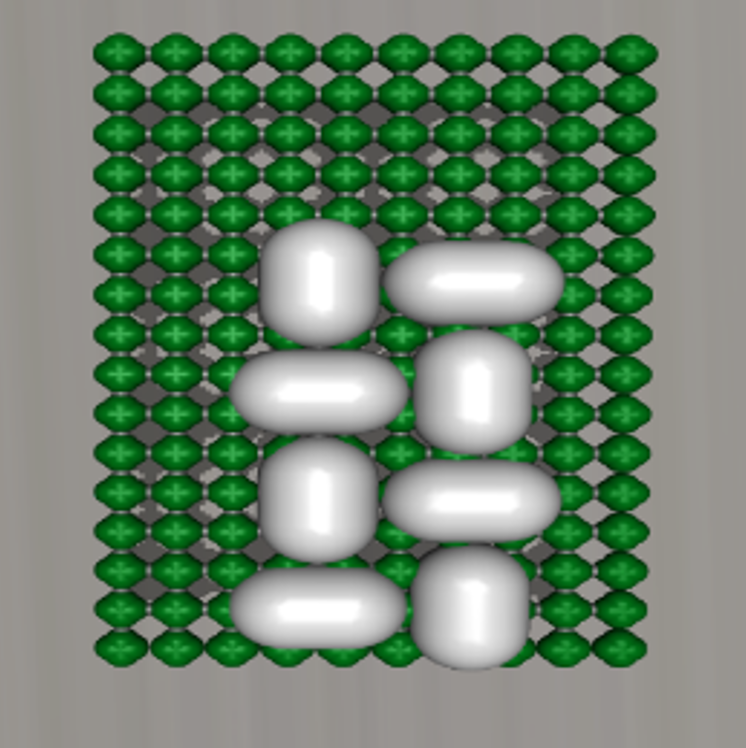}
      \caption{coin model}
    \end{subfigure}
    \begin{subfigure}{0.20\textwidth}
      \centering
      \includegraphics[width=\linewidth]{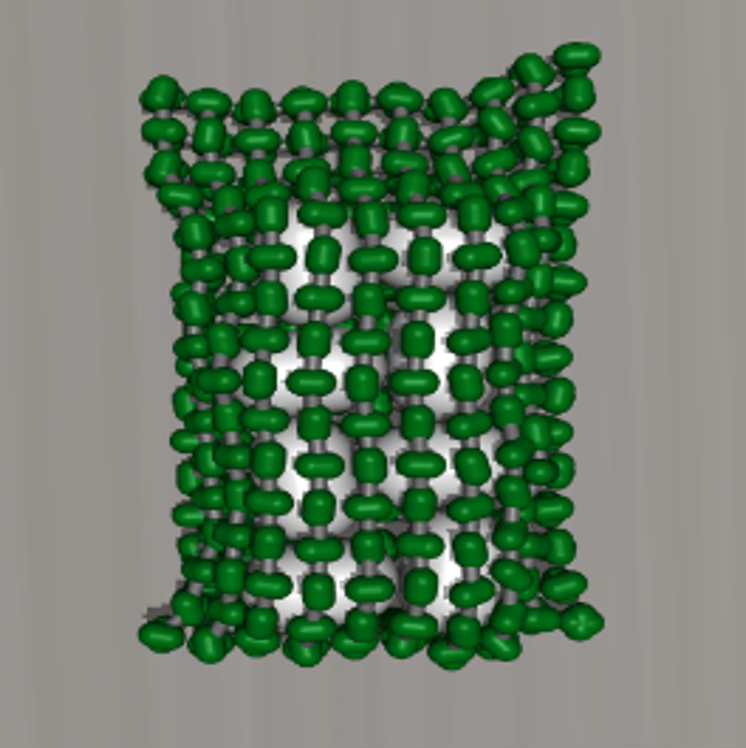}
      \caption{bag model}
    \end{subfigure}
    \caption{Simulations}
    \label{fig:simulation}
\end{figure}

\section{Boosted Motion Planning}
This section describes the boosted motion planning system for reducing the planning time of the robot manipulator by selecting a trajectory from a set of preplanned supporting trajectories and its subsequent correction. There are two modes of the boosted motion planning system: the supporting trajectories planning mode and the operation mode.

\emph{Supporting trajectories planning mode.} The operation space of the robot has been divided to some cells which are called "areas of interest". Then, the initial position of the robot manipulator  (Figure \ref{fig:Areas of interest}-a) and the set of possible rotations of the robot end effector (Figure \ref{fig:Areas of interest}-b) corresponding to the cells are defined. For each cell center, forward and backward supporting trajectories are planned (Figure \ref{fig:supporting trajectory}) and all supporting trajectories are saved to the database.
\begin{figure}
    \centering
    \includegraphics[scale=0.42]{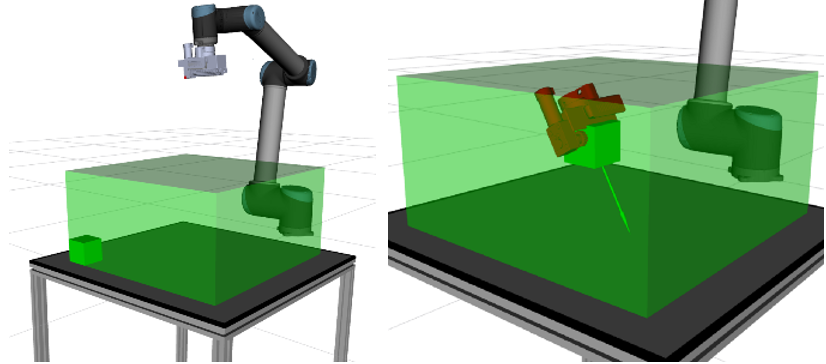}
    \caption{One of the cells and the initial position of the robot manipulator. b) Example of rotations from the set of desired rotations of the end effector.}
    \label{fig:Areas of interest}
\end{figure}
\begin{figure}
    \centering
    \includegraphics[scale=0.35]{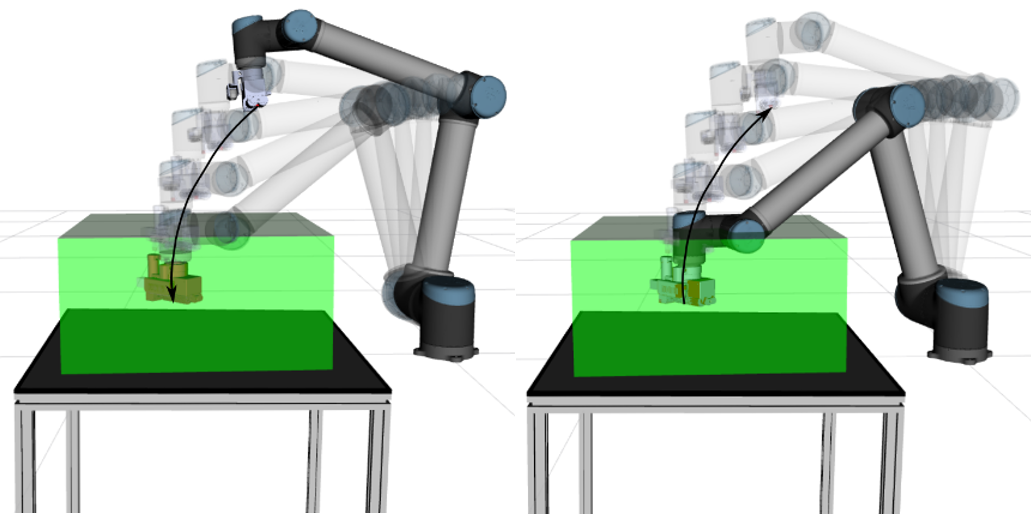}
    \caption{Left: Forward supporting trajectory. Right: Backward supporting trajectory}
    \label{fig:supporting trajectory}
\end{figure}
The configuration of areas of interest is shown in Figure \ref{fig:Areas of interest2}. Since bags can have an infinite number of pose variations inside the trolley, a greater number of different supporting trajectories with different target positions and orientations is needed to define areas of interest inside the trolley. Therefore, it contains 960 cells of size 55 mm and the set of 24 possible rotations of the robot end effector. Defining a large area of interest above the table (Figure \ref{fig:Areas of interest2}) is to transfer the grasping bag from above the trolley to above the table with a fixed trajectory. Due to this, the large area of interest above the table contains 16 cells of size 188 mm and has a set of 8 possible rotations of the robot end effector.

\begin{figure}
    \centering
    \includegraphics[scale=0.3]{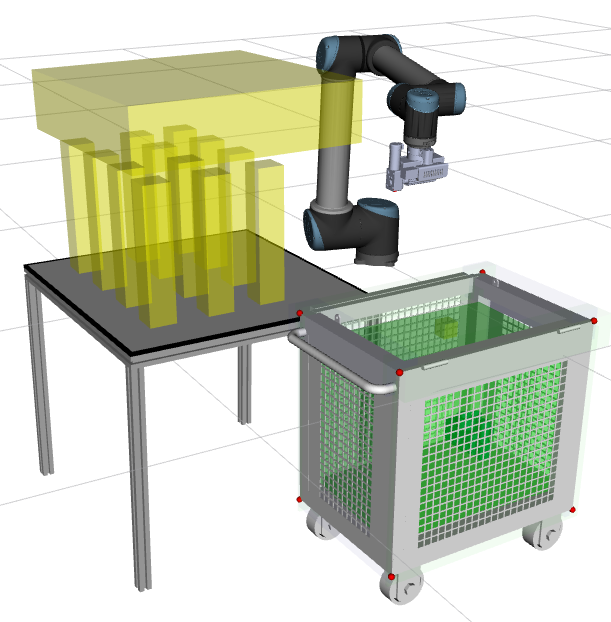}
    \caption{Areas of interest}
    \label{fig:Areas of interest2}
\end{figure}
To generate supporting trajectories, one of the most widely used planning algorithm based on discretization which is called Rapidly-exploring Random Trees (RRT) is chosen. A large number of RRT algorithm modifications have also been developed. For planning forward and backward trajectories for one cell and one desired rotation by MoveIt requires from 1 to 12 seconds. Thus, approximately 38,5 hours are necessary for planning supporting trajectories for the areas of interest in the trolley and 13 minutes are necessary for the large area of interest above the table. A large number of cells and the possible rotations of the robot end effector lead to a significant time planning of the supporting trajectories. To reduce the time of generating the supporting trajectories, it is possible to launch the planning process in several parallel threads.

\emph{Operation mode.} Algorithm 1 shows the sequence of steps of the system in this mode.

\begin{algorithm}[H]
 
 \textbf{Input}: Target pose of the robot end effector $pose_{target}$
 
 \textbf{Output}: Trajectory  \emph{trajectory}
 
 $\theta_{current}=getManipulatorJoints()$
 
 $pose_{current}=ForwardKinematic(\theta_{current} )$
 
 for each areaofinterest from areaofinterest:
 
        if  $pose_{target} in areaofinterest:$
        
              isProximity=checkProximity($pose_{current},pose_{initial}$ )
              
              if  isProximity:
              
                    $traj_{supporting}=ForwardTrajectoryFromDB(pose_{target} )$
                    
                    trajectory=correctTraj($traj_{supporting},pose_{target},\theta_{current}$ )
                    
                   end
                   
        else if  $pose_{current}$ in areaofinterest:
        
              isProximity=checkProximity($pose_{target},pose_{initial} $)
              
               if  isProximity:
               
                    $traj_{supporting}=BackwardTrajectoryFromDB(\theta_{current} )$
                    
                   trajectory=correctTraj($traj_{supporting},pose_{target},\theta_{current}$ )
                   
                  end
                  
     trajectory=planTrajectory($\theta_{current},pose_{target}$ )
     
     end 
 \caption{Boosted motion planning algorithms}
\end{algorithm}
In case when the target pose is inside the area of interest (Figure \ref{fig:supporting trajectory}-left), the system finds the trajectory at the last point of which the orientation of the robot end-effector is closest to the target orientation among all the forward trajectories recorded in the database. The system finds solution of the inverse kinematic for the target pose of the robot end-effector that satisfies the collision constraints and joint angles are closest to the robot joint angles at the end of the selected trajectory. The system adds the selected solution of the inverse kinematic problem at the end of the selected trajectory and the current robot manipulator joint angles at the beginning of the selected trajectory. In case when current pose is inside the area of interest (Figure \ref{fig:supporting trajectory}-right) the system finds the trajectory at the first point of which the robot manipulator joint angles are closest to the current robot manipulator joint angles among all the backward trajectories recorded in the database.
One operation, in which the robot manipulator picks one bag and places it on the table, has 6 trajectories. As noted earlier, MoveIt! requires 1-12 seconds for planning one trajectory. Boosted motion planning system requires 0.01-0.1 seconds for searching and modifying trajectories. Table \ref{comparison} shows a quantitative comparison of time to find a trajectory using RRT with MoveIt! and our approach for different number of bags. The results show that using our approach can significantly reduce the planning time and make the robot operation faster.

\begin{table}[h!]
\centering
\caption{Quantitative comparison of boosted motion planning and RRT}
\begin{tabular}{ c| c| c }
 Number of bags &  RRT from MoveIt! &   Boosted motion planning   \\
  \hline
 1  & 4.67 s     & 0.16 s \\
 5 & 25.21  s     & 0.92 s  \\
 20 & 88.73 s &  2.54 s \\
\end{tabular}

\label{comparison}
\end{table}

\section{Machine Teaching and Curriculum Learning}

Project Bonsai is a cloud-based solution offered by Microsoft that leverages machine teaching and reinforcement learning to solve autonomous systems problems. At its core, Bonsai leverages curriculum learning for training Bonsai brains in a sequential lesson-based fashion, allowing a subject-matter-expert to infuse her design on the brain’s learning strategy in the form “scenarios”, which control the level of domain randomization, episode configuration, as well as other aspects of the simulation environment, in order to improve efficiency of learning and safety at time of deployment. Each Bonsai brain is trained using optimized and scalable implementations of distributed reinforcement learning algorithms. Moreover, Bonsai allows for massive simulator scaling via asynchronous workers running on Azure containers.

For the purposes of training and deploying the coinbot brain, we utilized the Bonsai platform for training a continuous-action system using policy gradient algorithms. We trained our brains using Bonsai implementations of both SAC \cite{Haarnoja2018SoftAA} and PPO \cite{Schulman2017ProximalPO} using the MDP formulation described below.
We segment our state observations into three distinct groups:

\begin{enumerate}
  \item Feature points are extracted from the observed bag surface as discussed in the perception section. These are five 3d points, therefore, there are 15 values of representation.
  \item Relative position of the bags inside the trolley is represented by categorical value. All possible bag positions and their labels has been shown on Figure \ref{fig:bag-positions}.
  \item Orientation of the bag which is described by a single value that it equals 1 if the bag is placed in a vertical form and it is 0 for other situations.
\end{enumerate}

\textbf{Terminals} are Conditions which define the completion of a given episode and the start of another. An example of an obvious episode termination is successful grasping and lifting of the bag. For the experiments described in this paper we selected five iterations for the episode length, to allow the brain to make corrective adjustments if it is not capable of solving the grasping task. This leads to a simplified terminal condition: gripper returns to initial position with a newly initialized bag position if the number of attempts exceeds 6 or if the grasping task was successful.

\textbf{Actions} are represented as 6 continues variables, which predict the target gripper pose (position and orientation) for grasping. They are normalized according to the predicted bag bounding box by the object detection stage, and are represented as relative pose of the gripper inside this bounding box.

\textbf{Reward} is a measure of our agent’s behavior and are calculated at the iteration-level. We decomposed the reward function in few components: reward of the achievable pose (value 1.0), reward of the grasping (value 2.0) and reward of the pulling (value 5.0). To facilitate grasping on first iteration we also add discounting factor, which is presented in Table \ref{Discounted factor}.

\begin{table}[h!]
\centering
\caption{Discounted factor}
\begin{tabular}{ c| c| c |c|c|c|c}
 Iter. num &  0 & 1 & 2 & 3 & 4 & 5  \\
  \hline
 discounter  & 1.0 & 0.55 & 0.3 & 0.15 & 0.08 & 0.04 \\
\end{tabular}

\label{Discounted factor}
\end{table}

\subsection{Domain Randomization and Lessons}

The simplest scenario for RL in this task is a bag laying flat on the bottom of the trolley. Although it is easy to learn but it is not generalized neither to bags which are near sides of the trolley nor to bags are placed at upper heights. Therefore, we have presented several options for placing bags inside the cart, which it is called hardness-levels (\ref{fig:bag-positions}). The hardness-level's label contains the numbers of the cart walls, near which the corresponding bag initialization areas are located. For example, 24 indicates the area at the corner between walls 2 and 4 of the cart. So, we can manage the process of randomizing the initial positions of bags in strictly defined areas inside the cart.  Using the Bonsai platform, we defined a teaching curriculum\footnote{\url{https://docs.microsoft.com/en-us/bonsai/inkling/keywords/lesson}}  where the initial positions of the bags are randomly sampled from a uniform discrete distribution as shown in Figure \ref{fig:bag-positions}, and depths sampled continuously over a range $[20 , 63]$ Cm related to the base of the robot. During training, the Bonsai platform adapts the sampling distribution for new episode configurations gradually based on the current test-time capabilities of the learned brains, following the approach in \cite{Portelas2019TeacherAF}. The sampling scheme is learned through a Gaussian Mixture Model for a teacher algorithm that learns to generate a learning curriculum based on current Learning Progress of the policy.

\begin{figure}
    \centering
    \includegraphics[scale=0.32]{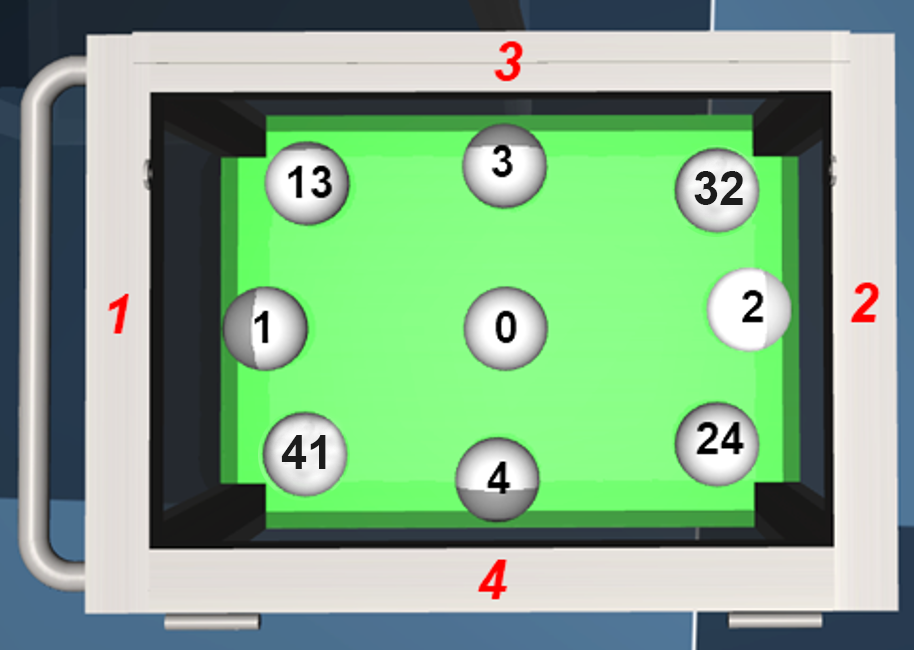}
    \caption{Bag Positions-Hardness level}
    \label{fig:bag-positions}
\end{figure}

\section{Experiment and Results}

Our top performing trained policy was transferred to a physical UR10 robotic arm. This policy achieves a test success rate of over 96\% on various bag positions and heights in trolley. Learning curves and distribution of success rates of the trained results are shown in Figure \ref{results-graphic}.

\begin{figure}

    \centering
      \includegraphics[scale=0.45]{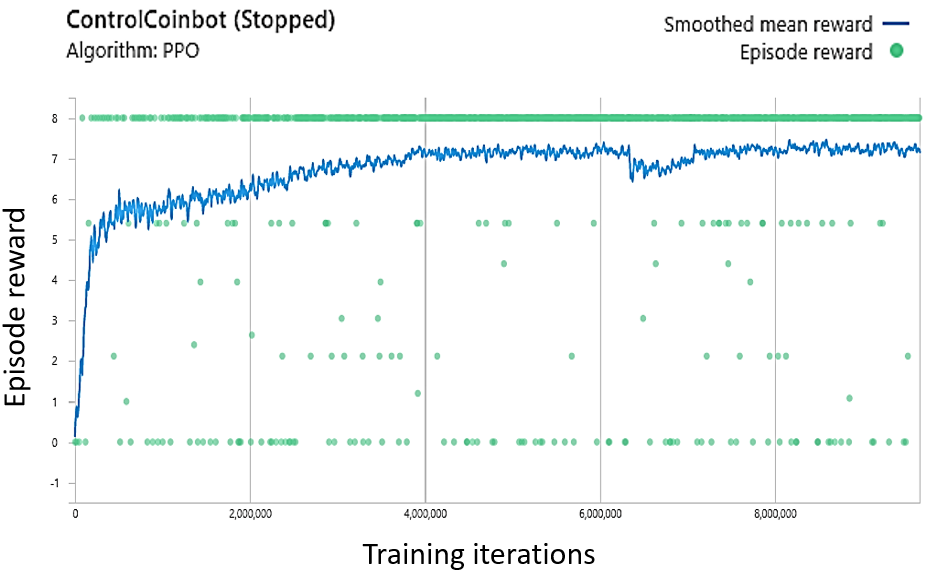}
    \caption{Learning Curves}
    \label{results-graphic}
\end{figure}

\begin{figure}

    \centering
      \includegraphics[width=\linewidth]{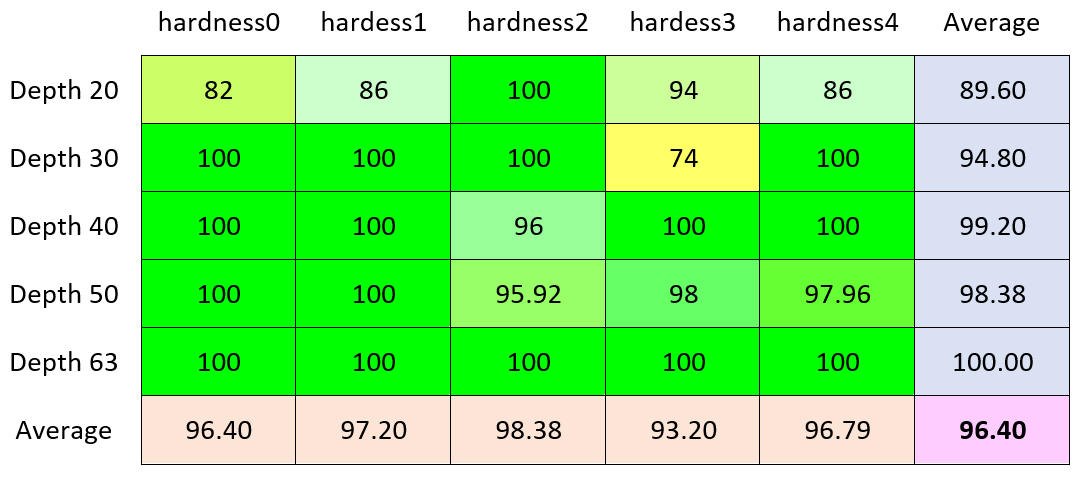}
    \caption{Results}
\end{figure}

Episode rollouts were collected from a distributed collection of 400 simulators running in Azure Batch, where for each simulator the initial position of the bag was generated randomly at all hardness levels inside the cart at heights -63cm, -40cm, -20cm in relation to the base point of the robot. The top performing policy was achieved for a brain trained with PPO and using lessons configured through domain randomization over 4 unique scenarios, and showed more than 93\% average success in pulling the bag at all hardness positions and more than 89\% average success at all depths. The minimum success occured at the highest level in the cart.

To simulate the trolley being filled with bags, the bottom of the trolley was raised to heights -63 cm, -40 cm, -20 cm related to the base of the robot. While there was only a single bag in the trolley during simulation, in real experiment, a large number of bags were placed in the trolley. To assess the quality of the model on a real robot, 19 bags of different weights were used. The weight distribution of the bags was as follows: 5 bags of 1 kg, 2 bags of 1.5 kg, 8 bags of 2 kg, 3 bags of 3 kg, 1 bag of 4.5 kg. In real-world test trials, the top policy brain was able to remove all 19 bags from the trolley without human assistance in 98\% of the all experiments. Table \ref{first attempt} shows the percentage of different situations for the first attempt to take the bag in different heights inside the cart. In the supplemental video\footnote{\url{https://www.youtube.com/watch?v=5o2Fqkbd8gc&feature=youtu.be}} we demonstrate the efficiency of our approach for unloading the coin bags with artificial brain and boosted motion planner.

\begin{table}[h!]
\centering
\caption{First attempt grasping results on different heights in real experiment.}
\scalebox{0.95}{
\begin{tabular}{ p{7mm}|p{8mm}|p{15mm}|p{12mm}|p{15mm}|p{8mm}}
 height \newline (cm) &  Success \newline (\%)  & Unsuccess \newline positioning(\%)& Error of \newline gripping(\%) & Not \newline achievable(\%) & Total \newline attempts  \\
  \hline
 -20  & 73.33 & 10.66 & 2.66 & 13.33 & 75 \\
 -40  & 76.47 & 10.58 & 0 & 9.41 & 85 \\
 -63  & 81.25 & 14.06 & 3.12 & 1.56 & 64 \\
\end{tabular}
}
\label{first attempt}
\end{table}

\section{Conclusion}

A new application of collaborative robots in cash center of the bank using artificial intelligence has been presented in this paper. A Bonsai brain has been trained to suggest the best configuration of the gripper for successful grasping and the boosted motion planner has been presented for fast and reliable operation. A patented gripper is introduced for this application which it can be extended for any other similar applications. Simulating complex environment for deep reinforcement learning and transferring the cloud brain which is trained on simulation to the local one and implementation of that on the real robot, has been done in this project. The output automation of heavy bags unloading, eradicates labors and increase the speed of sorting process in cash centers. The presented approach can be extended to other similar applications in warehouses and other similar environments. For future study, improvements to the perception system to detect human operators for collision avoidance will be studied, as well as generalizations to new bags and new trolleys.

\bibliographystyle{IEEEtran}
\bibliography{IEEEabrv,bibliography}

\end{document}